\documentclass[a4paper,review]{cas-sc}
\usepackage[authoryear,longnamesfirst]{natbib}

\def\tsc#1{\csdef{#1}{\textsc{\lowercase{#1}}\xspace}}
\tsc{WGM}
\tsc{QE}
\tsc{EP}
\tsc{PMS}
\tsc{BEC}
\tsc{DE}

\begin{document}
\let\WriteBookmarks\relax
\def\floatpagepagefraction{1} 
\def\textpagefraction{.001} 
\shorttitle{LLM Enhanced Action Recognition}
\shortauthors{Ruosi Wang et~al.}


\title [mode = title]{LLM Enhanced Action Recognition via Hierarchical Global-Local Skeleton-Language Model}                      

\author[1]{Ruosi Wang}
\ead{wangruosi@mail.nwpu.edu.cn}

\credit{Conceptualization, Investigation, Methodology, Writing – original draft}

\affiliation[1]{organization={Northwestern Polytechnical University},
                addressline={No.1 Dongxiang Road, Chang'an District }, 
                city={Xi'an},
                postcode={710129}, 
                state={Shaanxi},
                country={China}}
\author[1]{Fangwei Zuo}
\ead{fangweizuo@mail.nwpu.edu.cn}
\credit{Data curation, Software, Validation, Visualization, Writing – original draft}

\author[2]{Lei Li}
\ead{lilei@sdufe.edu.cn}
\credit{Investigation, Formal analysis, Data curation, Validation}

\affiliation[2]{organization={Shandong University of Finance and Economics},
                addressline={No.40 Shungeng Road, Licheng District}, 
                city={Jinan},
                postcode={Jinan}, 
                state={Shandong},
                country={China}}

\author[1]{Zhaoqiang Xia}[
                        orcid=0000-0003-0630-3339]
\credit{Funding acquisition, Investigation, Project administration, Supervision, Writing – review \& editing}

\cormark[1]
\ead{zxia@nwpu.edu.cn}

\cortext[cor1]{Corresponding author}

\begin{abstract}
Skeleton-based human action recognition has achieved remarkable progress in recent years. However, most existing GCN-based methods rely on short-range motion topologies, which not only struggle to capture long-range joint dependencies and complex temporal dynamics but also limit cross-modal semantic alignment and understanding due to insufficient modeling of action semantics. To address these challenges, we propose a hierarchical global-local skeleton-language model (HocSLM), enabling the large action model be more representative of action semantics. First, we design a hierarchical global-local network (HGLNet) that consists of a composite-topology spatial module and a dual-path hierarchical temporal module. By synergistically integrating multi-level global and local modules, HGLNet achieves dynamically collaborative modeling at both global and local scales while preserving prior knowledge of human physical structure, significantly enhancing the model’s representation of complex spatio-temporal relationships. Then, a large vision-language model (VLM) is employed to generate textual descriptions by passing the original RGB video sequences to this model, providing the rich action semantics for further training the skeleton-language model. Furthermore, we introduce a skeleton-language sequential fusion module by combining the features from HGLNet and the generated descriptions, which utilizes a skeleton-language model (SLM) for aligning skeletal spatio-temporal features and textual action descriptions precisely within a unified semantic space. The SLM model could significantly enhance the HGLNet’s semantic discrimination capabilities and cross-modal understanding abilities. Extensive experiments demonstrate that the proposed HocSLM achieves the state-of-the-art performance on three mainstream benchmark datasets: NTU RGB+D 60, NTU RGB+D 120, and Northwestern-UCLA.
\end{abstract}


\maketitle

\section{Introduction}
In recent years, with the rapid development of emerging fields such as autonomous driving, healthcare, video surveillance, and human-computer interaction, computer vision has become a central focus of research in intelligent systems. As one of the fundamental tasks in visual understanding, Human Action Recognition (HAR) aims to automatically analyze and classify human action categories or behavioral states from videos, depth images, skeletal sequences, or other modalities. It serves as a critical component for achieving high-level semantic understanding. This task has been widely applied in areas including emotion recognition, behavior analysis, and sports assessment, playing an essential role in enhancing the intelligence and safety of systems (\cite{3,song2026trajectory}).

Early action recognition methods primarily relied on RGB images and optical flow features (\cite{6}), which extract motion patterns by analyzing color information and trajectories across video frames. However, these modalities are highly sensitive to viewpoint variations, illumination changes, background clutter, and occlusions (\cite{7}), leading to unstable performance in complex real-world scenarios. Compared to RGB images and optical flow features, skeleton sequences provide a more robust and stable representation of actions, focusing on the spatiotemporal changes of human joints, overcoming challenges posed by viewpoint variations and complex backgrounds. Skeleton data exhibits stronger invariance to viewpoint changes and can effectively handle occlusion issues, thereby improving the accuracy and robustness of action recognition. Consequently, an increasing number of studies (\cite{4}) are shifting towards skeleton-based action recognition to overcome the limitations of traditional methods in complex environments. By focusing on key joints and motion trajectories of the human body, skeleton sequences not only reduce the interference caused by environmental changes but also maintain high accuracy and stability in various complex scenarios. Additionally, the low computational complexity of skeleton data makes it more advantageous in real-time applications, further driving its widespread use in the field of action recognition.

In the field of skeleton-based action recognition, traditional methods mainly rely on handcrafted feature extraction and classical machine learning models. Handcrafted feature extraction typically converts skeleton data into feature vectors such as joint positions, angle changes, velocities, and accelerations. These features can reflect the spatio-temporal changes of actions to some extent, but due to their reliance on manual design, they often fail to capture the high-level semantic information of complex actions comprehensively, especially in high-dimensional data and dynamic complex scenes. Classical machine learning models, such as Support Vector Machines (SVM) (\cite{67}) and Hidden Markov Models (HMM) (\cite{68}), divide action categories by constructing optimal hyperplanes and describe the temporal dynamics of action sequences, respectively. While these methods achieve effective results in simpler scenarios, they still have significant limitations in handling complex spatio-temporal patterns and long-range dependencies. With the development of deep learning techniques, skeleton-based action recognition methods have gradually shifted towards using deep neural networks, particularly Convolutional Neural Networks (CNNs) (\cite{9}), Long Short-Term Memories (LSTMs) (\cite{12}), and Recurrent Neural Networks (RNNs) (\cite{shaikh2024cnns}). These methods can automatically learn discriminative feature representations, enabling more effective modeling of complex spatio-temporal patterns and long-range dependencies, which significantly improves action recognition performance. However, due to the inherent graph structure of skeleton data, these methods struggle to fully model the non-Euclidean structural relationships between joints. Furthermore, CNNs, RNN, and LSTMs still have limitations in capturing long-term temporal dependencies, especially when handling complex action sequences and long time spans, as they face the problem of gradient vanishing.

To better leverage the topological structure of skeletons, Graph Convolutional Networks (GCNs) (\cite{16}) were introduced for this task. GCNs can naturally model the graph-structured relationships between joints and bones while capturing both spatial and temporal dynamics, thereby significantly outperforming traditional methods in terms of recognition accuracy and robustness. Existing methods primarily encompass typical paradigms such as spatio-temporal graph convolution, multi-scale modeling, dynamic graph convolutions, and GCN-based Transformer. Spatio-temporal graph convolution methods (\cite{18,19}) construct explicit spatio-temporal graphs to effectively model joint motion patterns. However, they rely on predefined static adjacency matrices, which struggle to adapt to the highly dynamic inter-joint interaction patterns that vary across different actions. To address the limitations of static graph topologies, multi-scale modeling approaches (\cite{20,21}) expand the spatio-temporal receptive field to enhance feature representation capability. Despite these improvements, their topological structures remain relatively rigid, limiting their effectiveness in capturing long-range temporal dependencies and action-specific dynamic joint associations. Dynamic graph convolutional methods (\cite{22,23}) demonstrate strong performance in complex action and long-sequence tasks by adaptively adjusting the graph structure. The GCN-based Transformer method (\cite{69,70}) combines the spatial graph structure modeling capability of GCNs with the powerful time series modeling ability of Transformers. By dynamically adjusting the relationships between skeleton joints through the self-attention mechanism, it overcomes the limitations of traditional methods in modeling complex behaviors. Nevertheless, these methods still face challenges in simultaneously modeling the dynamic evolution of global and local features, and they generally lack sufficient capability to capture high-level semantic information inherent in complex human actions.

Despite the remarkable progress achieved by the aforementioned methods in skeleton-based action recognition, several critical limitations remain unresolved. On one hand, the predominant focus on geometric and kinematic patterns results in insufficient semantic understanding, rendering these models less effective for fine-grained actions and cross-scenario generalization. On the other hand, reliance on purely visual modalities without effective alignment to linguistic information severely restricts model interpretability and higher-order cognitive capabilities. To address these challenges, we propose a \textbf{H}ierarchical gl\textbf{o}bal-lo\textbf{c}al \textbf{S}keleton-\textbf{L}anguage \textbf{M}odel (shortly, \textbf{HocSLM}). First, the Hierarchical Global-Local Network (HGLNet) is designed to effectively capture both spatial and temporal dynamics in skeleton data, overcoming the shortcomings of traditional approaches in modeling long-range temporal dependencies and dynamic joint interactions. Second, inspired by the recent development of large language model (LLM), we further employ a large vision-language model (VLM) to generate textual descriptions by passing the original RGB video sequences to this model, providing rich action semantics for training the skeleton-language model. Third, the Skeleton-language Sequential Fusion module (SSF) is incorporated, which establishes cross-modal alignment between skeleton sequences and textual action descriptions via a skeleton-language alignment mechanism, thereby significantly enhancing semantic comprehension and generalization ability. The main contributions of this work are summarized as follows:
\begin{itemize} 
\item We propose a novel large action model (HocSLM) for skeleton-based action recognition, which exploits the LLM to enhance the motion representation, providing the enriched text descriptions to describe the action semantics.
\item We design a skeleton network (HGLNet) that effectively mitigates the limitations of traditional spatio-temporal modeling of actions in capturing long-range temporal dependencies and dynamic inter-joint interactions.
\item We introduce a skeleton-language fusion module (SSF), which significantly improves the model's semantic discrimination and task generalization capabilities through cross-modal vision-language alignment.
\item Extensive experiments on NTU RGB+D 60, NTU RGB+D 120, and Northwestern-UCLA datasets are performed and consistently show that the proposed HocSLM surpasses the state-of-the-art methods across all standard evaluation protocols.
\end{itemize}

The rest of this paper is organized as follows: Section \ref{sec_rw} provides a brief review of related works and existing methods. Section \ref{sec_method} presents the proposed HocSLM in detail, including its network architecture and key module designs. Section \ref{sec_experiment} reports experimental setups and results, comparing our method with state-of-the-art approaches. Finally, Section \ref{sec_conclusion} summarizes the entire paper and discusses future research directions.

\section{Related Work}
\label{sec_rw}
\subsection{Skeleton-Based Action Recognition}
Early skeleton-based action recognition methods relied on hand-crafted geometric and kinematic features (\cite{24}), which suffered from high design complexity, poor noise robustness, and limited capability to comprehensively capture essential factors of human motion. With the advent of deep learning, CNNs (\cite{9}) and RNNs (\cite{shaikh2024cnns}) gradually replaced hand-crafted features and enabled end-to-end training. However, neither paradigm explicitly modeled the inherent graph structure of skeletons, which significantly constrained their ability to recognize complex actions.

To overcome the limitations of conventional approaches, Graph Convolutional Networks (GCNs) were introduced into skeleton-based action recognition. GCNs can naturally align with the topological structure of human joints and effectively model the spatial relationships between joints. Building upon this foundation, the pioneering work ST-GCN (\cite{18}) was the first to extend graph convolutions to the spatio-temporal dimension. By constructing spatio-temporal graphs, it simultaneously captures spatial correlations between joints and temporal dynamics of motion evolution. However, this approach relies on predefined fixed adjacency matrices, making it difficult to adapt to the dynamic changes in joint interactions across different actions. 

To achieve stronger multi-scale spatio-temporal modeling, MS-G3DNet (\cite{20}) was proposed to disentangle spatial and temporal dimensions by combining multi-order adjacency matrices with multi-branch temporal convolutions and 3D convolutions. Despite its effectiveness, MS-G3DNet still heavily depends on predefined topologies and struggles to adapt to the dynamic evolution of joint interactions. To address these issues, self-attention mechanisms have been incorporated into skeleton-based action recognition. InfoGCN by \cite{26} integrated the information bottleneck principle with attention mechanisms to capture context-aware skeleton topologies, significantly improving both accuracy and robustness. Similarly, 2s-AGCN by \cite{27} employed adaptive graph convolution modules within a two-stream framework to enable dynamic learning of spatial topologies and multi-feature fusion. Nevertheless, these methods still suffer from insufficient modeling of long-range temporal dependencies and limited optimization of dynamic topologies during complex actions. More recent advances have focused on fully adaptive graph modeling. Dynamic-GCN proposed by \cite{28} introduced a learnable dynamic graph generation module that enables end-to-end adaptive optimization of spatio-temporal topologies, demonstrating superior representation capacity on complex action sequences. CTR-GCN proposed by \cite{29} further enhances channel-wise dynamic topology learning, yielding stronger extraction of diverse action features. Meanwhile, Shift-GCN proposed by \cite{22} leveraged shift graph convolutions and learnable adjacency matrices to achieve efficient and flexible spatio-temporal feature aggregation. Based on this foundation, \cite{61} proposed the Dynamic Group Spatio-Temporal Graph Convolutional Network (DG-STGCN), which consists of DG-GCN for dynamic spatial modeling using learned affinity matrices and DG-TCN for adaptive temporal modeling with group convolutions and varying receptive fields. To more effectively capture critical structural features and long-range dependencies within graphs, HD-GCN proposed by \cite{62} effectively extracted key structural adjacencies and long-range edges by introducing hierarchical decomposition graphs and attention-guided hierarchical aggregation. \cite{63} proposed Decoupled Static-Dynamic Co-occurrence Graph Convolution (DSDC-GConv), which overcomes the limitations of static skeleton topology in capturing joint discriminative relationships and modeling dynamic motion information by decomposing inter-frame and intra-frame joint dependencies to refine and adapt the graph topology. To further overcome the limitations of traditional methods in modeling complex actions, GCN-based Transformer methods have emerged in recent years. The proposed LG-STFormer by \cite{69} demonstrates strong capabilities in capturing both spatial and temporal features of skeletons by combining local and global attention modules. TranSkeleton by \cite{70} effectively unifies spatial and temporal modeling of skeleton sequences through partition-aggregation temporal Transformer and topology-aware spatial Transformer, significantly enhancing the capture of long-range dependencies and subtle temporal structures.

Although existing GCN based skeleton action recognition methods have made progress in graph structure and spatio-temporal modeling, they still face challenges such as high computational complexity and insufficient modeling of long-range dependencies. To address these issues, we propose HGLNet for extracting visual features, which integrates the collaborative architecture of CTS and DHT. Compared to methods such as InfoGCN (\cite{26}) and Dynamic-GCN (\cite{28}), CTS dynamically models spatial features from both global and local perspectives through a dual-branch design, accurately capturing joint relationships at different scales, while InfoGCN primarily relies on graph convolution and attention mechanisms for static spatial modeling, failing to capture the multi-scale variations in spatial features; DHT combines long-range and short-range temporal dependencies, significantly enhancing the modeling of temporal features and optimizing the capture of complex temporal dynamics, whereas Dynamic-GCN, although it optimizes the spatio-temporal topology through dynamic graphs, still has limitations in modeling long-range dependencies and dynamic temporal sequences. Additionally, the dynamic weighted fusion mechanism and physical connection priors introduced by CTS and DHT effectively integrate global and local information, improving accuracy and robustness while reducing computational overhead, which has not been fully addressed in existing methods.

\subsection{Large Language Model for Action Recognition}
In recent years, VLMs have achieved remarkable progress in multimodal learning tasks. CLIP by \cite{32} and ALIGN by \cite{33} employ contrastive learning to map visual and linguistic modalities into a unified semantic space, establishing a foundation for cross-modal understanding. UniCL proposed by \cite{34} constructs a unified contrastive learning framework that reorganizes image-label pairs into a joint image-text-label representation. ActionCLIP by \cite{35} extends this paradigm to video-based action recognition and proposes a new ``pre-train, prompt, and fine-tune'' paradigm.

With significant advancements in VLMs, researchers have begun incorporating LLMs into the field of multimodal learning. LLMs (\cite{36}) possess robust semantic understanding and contextual reasoning capabilities, enabling exceptional performance in multimodal tasks. By integrating visual modalities with linguistic information, LLMs better capture spatio-temporal dynamics and fine-grained actions. Flow4Agent by \cite{64} enhances LLM performance in video action recognition by incorporating optical flow as a motion prior, enabling the capture of fine-grained dynamics. VideoPerceiver by \cite{65} optimizes spatio-temporal perception modules, enabling LLMs to process short-duration and transient events more efficiently, thereby enhancing fine-grained action recognition. STORM by \cite{66} further improves LLM recognition capabilities in long videos and complex scenes through spatio-temporal token dimensionality reduction, contrastive learning, and temporal modeling. However, while these approaches advance video action understanding, they remain constrained by reliance on video-based fine-grained motion information. In contrast, LLMs demonstrate potent knowledge transfer capabilities in skeleton-based behavior recognition. GAP-Net by \cite{38} innovatively leverages LLMs to generate fine-grained action descriptions, introducing linguistic prior knowledge into visual representation learning. To overcome the limitations of static semantics in temporal modeling, LLM-AR by \cite{39} directly recognizes skeleton sequences using internal LLM knowledge, demonstrating the potential of LLM action understanding. However, fine-grained visual-linguistic alignment remains insufficient. Furthermore, existing methods such as DVTA by \cite{40} and EMP by \cite{41} mostly rely on text encoders during inference, significantly increasing computational overhead and deployment complexity. LA-GCN proposed by \cite{30} enhances skeleton feature learning by integrating prior knowledge from LLM. However, its reliance on static prior knowledge limits its performance in handling dynamic spatio-temporal dependencies and complex action sequences.

To address the issues of insufficient fine-grained alignment and low inference efficiency, we propose SSF to align the skeleton and semantics. This module achieves efficient alignment and semantic enhancement between the model's internal spatio-temporal skeleton features and textual descriptions without relying on external text encoders, thereby improving recognition accuracy while significantly reducing computational overhead and deployment complexity.

\section{Methodology}
\label{sec_method}
\subsection{The Overall Architecture}
\begin{figure}[pos=htbp]
    \centering
    \includegraphics[width=0.95\linewidth]{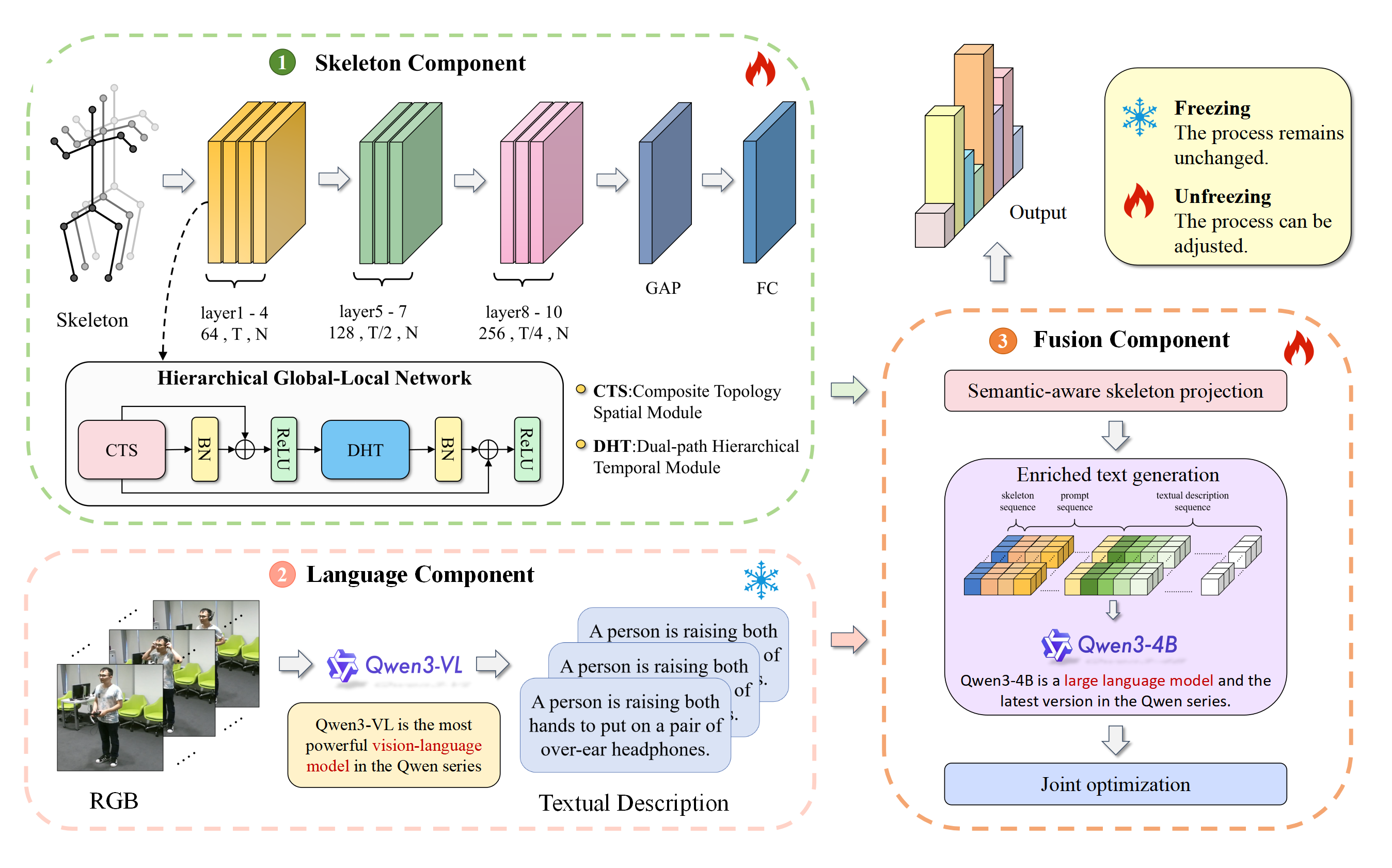}
    \caption{The architecture of the proposed HocSLM, including 1) Skeleton Component, 2) Language Component, and 3) Fusion Component.}
    \label{fig_HocSLM}
\end{figure}

For the skeleton-based action recognition task, we propose HocSLM, as shown in Figure \ref{fig_HocSLM}. The model takes skeleton sequences and textual description sequences as input, significantly enhancing the representation ability of skeleton-based action recognition through hierarchical spatiotemporal modeling and semantic space alignment. Specifically, HocSLM consists of three main components: the \textit{skeleton component}, the \textit{language component}, and the \textit{SSF component}. 

During the \textbf{training process}, the skeleton component first extracts skeleton features using the skeleton recognition model (HGLNet) and performs spatial and temporal modeling. Specifically, the CTS module captures spatial relationships between joints, while the DHT module models temporal dynamics. The extracted skeleton features are subsequently reduced in dimensionality through Global Average Pooling (GAP) and processed by a fully connected layer (FC) for action classification. The language component generates corresponding textual descriptions using a pre-trained large language model (Qwen3-VL). The SSF component then tokenizes these text descriptions and concatenates them with the skeleton features to form the skeleton-language input. This input is passed to a large language model (Qwen3-4B) for autoregressive text generation, ensuring the alignment of skeleton and text information, thereby enhancing both action recognition and text generation performance. Throughout this process, the model is jointly optimized to minimize both text generation loss and classification loss, ensuring accurate action recognition while generating corresponding textual descriptions, thereby improving semantic alignment and overall performance. 

In the \textbf{inference phase}, the model utilizes the learned skeleton feature extraction and text generation capabilities to generate image descriptions. First, the skeleton component extracts features and performs modeling, followed by GAP reduction and FC processing, generating skeleton feature sequences. Next, the language component combines the extracted skeleton features with a fixed text prompt and inputs them into the Qwen3-4B model for descriptive text generation. During inference, the model no longer undergoes training but performs inference based on the learned weights, generating textual descriptions corresponding to the image content and completing the task of automatic understanding and generation of image content.

\subsection{Composite-topology Spatial Module}
\begin{figure}[pos=htbp]
    \centering 
    \includegraphics[width=0.9\linewidth]{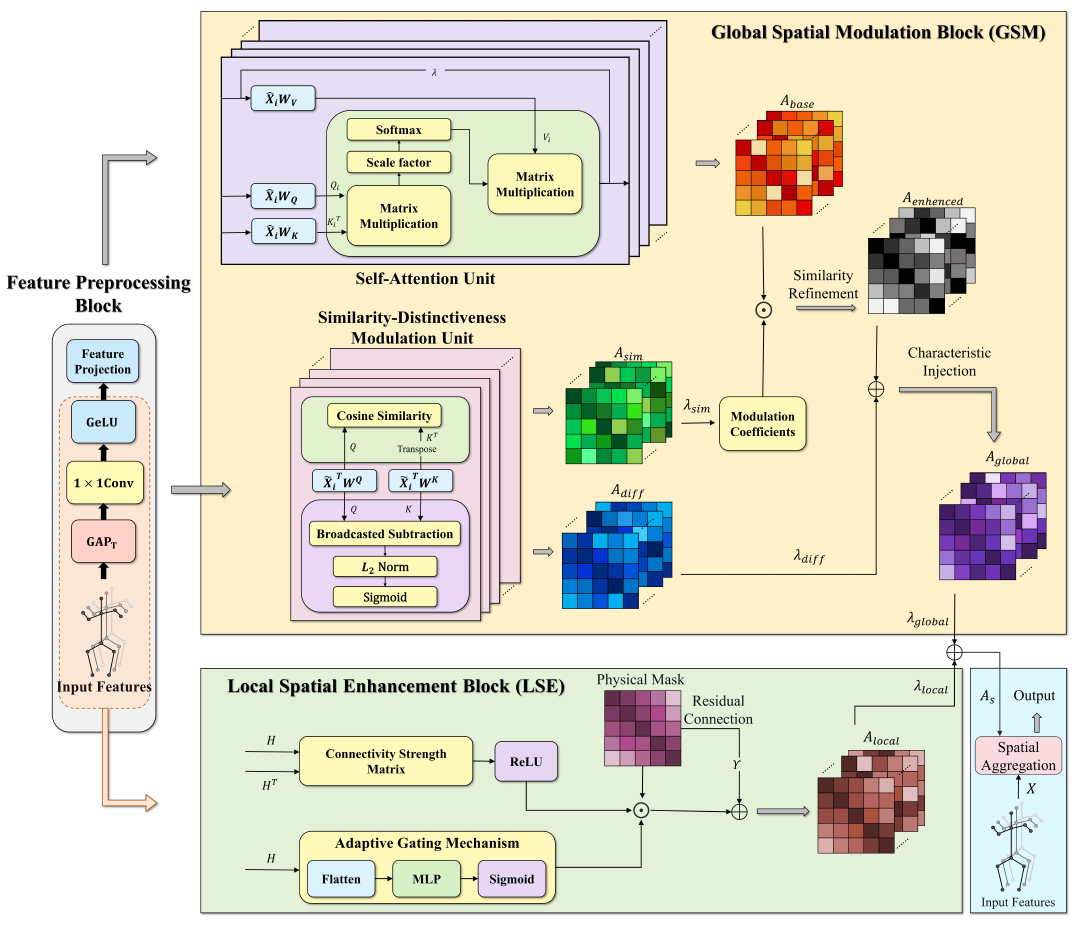}
    \caption{Overall architecture of the CTS. The leftmost module is the feature preprocessing block. The upper orange region corresponds to the GSM, and the lower green region corresponds to the LSE.}
    \label{fig_CTS}
\end{figure}

To address the limitations of existing skeleton-based action recognition methods in spatial topology modeling (\cite{18}) and their insufficient utilization of human structural priors (\cite{42}), we propose the Composite-Topology Spatial module (CTS) in Figure \ref{fig_CTS}. The module employs a dual-branch design consisting of Global Spatial Modulation block (GSM) and Local Spatial Enhancement block (LSE). This architecture preserves the rationality of physical connectivity while enabling dynamic modeling of joint collaboration patterns under different action modes, thereby achieving more robust and accurate spatial feature extraction.

The feature preprocessing block first aggregates temporal information through global average pooling along the temporal dimension while retaining the spatial dimension, thereby generating initial node-level features. Subsequently, these features are processed through the feature transformation layer and linear projection layer, and then map node representations into a high-dimensional space. The resulting output feature tensor is formulated as Equation \ref{eq_X1}.
\begin{equation}
\label{eq_X1}
\begin{split}
\overline{X} &= \textit{GeLU}(\omega(\textit{GAP}_T(X_{\textit{s}})))  \\
\tilde{X} &= \textit{Linear}(\overline{X})
\end{split}
\tag{1}
\end{equation}
where \( \textit{GAP}_T \) represents global average pooling across the temporal dimension, which compresses the input feature \( X_{\text{s}} \in \mathbb{R}^{C_{\textit{in}} \times T \times N} \) along the time axis \( T \) to yield the aggregated spatio-temporal feature \( \textit{GAP}_T(X_{\text{s}}) \in \mathbb{R}^{C_{\textit{in}} \times 1 \times N} \). The linear transformation function \( \omega(\cdot) \) modifies the channel dimension \( C_{\textit{in}} \) to \( C_{\textit{in}} / r_{\tau} \) to reduce computational complexity. The \( \text{Linear}(\cdot) \) operation performs linear projection, mapping the node feature \( X \in \mathbb{R}^{C_{\textit{in}} / r_{\tau} \times N} \) to a high-dimensional space \( \mathbb{R}^{C_{\textit{in}} / r_{\tau} \times N \times N} \). For the first layer, $ X_{\text{s}}$ is the original skeleton sequence \( X_t \in \mathbb{R}^{N \times C} \), where \( N \) denotes the number of joints and \( C \) denotes the feature dimension per joint. The output tensor contains pairwise associations between all \( N \times N \) node pairs, providing the network with associative information across nodes.

\subsubsection{Global Spatial Modulation Module}
To overcome the limitations of traditional GCNs in modeling complex spatial dependencies between joints in skeleton sequences, we propose the GSM based on self-attention mechanisms. This module introduces a similarity-distinctiveness modulation block to enhance the spatial representation capability of joint features from the dual perspectives of collaborative motion similarity and discriminative difference, and uses a hybrid modulation strategy to optimize global feature fusion. Furthermore, through cross-channel joint dynamic modeling, it achieves collaborative learning of shared topological structures and channel-specific correlations, thereby significantly improving performance and generalization capabilities in human action recognition tasks.

First, multi-head linear projection is performed on each input feature \( \tilde{X}_i \in \mathbb{R}^{N \times N} \) to generate query vector \( Q_i \), key vector \( K_i \), and value vector \( V_i \). Then, the similarity between joints is calculated using the self-attention mechanism to obtain the attention score. The correlation between the query vector and key vector is computed to derive the attention weight matrix, which is then weighted and summed with the value vector to produce the base adjacency matrix \( A^{i}_{\textit{base}} \). The specific process is expressed by Equation \ref{eq_Abase}.
\begin{equation}
\label{eq_Abase}
\begin{split}
Q_i &= W_Q \cdot \tilde{X}_i, \quad K_i = W_K \cdot \tilde{X}_i, \quad V_i = W_V \cdot \tilde{X}_i \\
A^{i}_{\textit{base}} &= \textit{Softmax} \left( \frac{Q_i K_i^T}{\sqrt{N}} \right) V_i
\end{split}
\tag{2}
\end{equation}
where \( W_Q, W_K, W_V \in \mathbb{R}^{N \times N} \) are parameter matrices shared by all channels. This process ensures that each channel can independently construct joint spatial dependency relationships and achieve dynamic aggregation of features.

In addition, we introduce a similarity-distinctiveness modulation block to further optimize the joint dependency modeling. The block enhances the model's expressive capability by computing the similarity matrix \( A^{i}_{\textit{sim}} \) and distinctiveness matrix \( A^{i}_{\textit{diff}} \) between joints, addressing synergistic motion and divergence characteristics respectively. The formulations are given in Equation \ref{eq_Asd}.
\begin{equation}
\label{eq_Asd}
\begin{split}
A^{i}_{\textit{sim}} &= \frac{\tilde{X}_i^Q (\tilde{X}_i^K)^T}{\left\| \tilde{X}_i^Q \right\| \left\| \tilde{X}_i^K \right\|} \\
A^{i}_{\textit{diff}} &= \textit{Sigmoid} \left(\left\| \tilde{X}_i^Q - \tilde{X}_i^K \right\|_2\right)
\end{split}
\tag{3}
\end{equation}
where \( \tilde{X}_i^Q \) denotes the query vector and \( \tilde{X}_i^K \) denotes the key vector. On this basis, a hybrid modulation method is designed to enhance key synergistic patterns and significant distinct features in the base features through multiplicative and additive operations, thereby constructing a global feature fusion matrix \( A^{i}_{\textit{global}} \) as shown in Equation \ref{eq_Agi}.
\begin{equation}
\label{eq_Agi}
A_{\textit{g}}^{i} = A_{\text{base}}^{i} \odot \left( 1 + \lambda_{\textit{sim}}^{i} A_{\textit{sim}}^{i} \right) + \lambda_{\textit{diff}}^{i} A_{\textit{diff}}^{i}
\tag{4}
\end{equation}
where \( \lambda_{\text{sim}} \) and \( \lambda_{\text{diff}} \) denote modulation coefficients, and \( \odot \) denote element-wise multiplication.

Finally, we construct the global feature representation by concatenating the enhanced features from each channel dimension and perform a linear transformation using the output weight matrix \( W^0 \in \mathbb{R}^{C_{\text{in}} / r \times C_{\text{out}}} \), resulting in the final global feature expansion matrix  \( A_{\textit{g}} \in \mathbb{R}^{C_{\text{out}} \times N \times N} \), as formulated in Equation \ref{eq_Ag}. 
\begin{equation}
\label{eq_Ag}
A_{\textit{g}} = \textit{Concat} \left(A_{\textit{g}}^1 + \lambda_1 \hat{X}_{\textit{g}}^1, A_{\textit{g}}^2 + \lambda_2 \hat{X}_{\textit{g}}^2, \dots, A_{\textit{g}}^{C_{\textit{in}} / r} + \lambda_{C_{\textit{in}} / r} \hat{X}_{\textit{g}}^{C_{\textit{in}} / r}\right) W^0
\tag{5}
\end{equation}
where \( \lambda \) denotes a residual weight parameter, and \textit{Concat} denotes concatenation along the channel dimension.

\subsubsection{Local Spatial Enhancement Module}
In above studies, it is observed that constructing channel-specific global spatial topologies for different output channels plays a critical role, significantly enhancing the model's ability to capture discriminative long-range dependencies. However, relying solely on global modeling tends to overfit on limited-scale datasets and impose weak explicit constraints on local physical connections, resulting in insufficient perception of local spatial patterns when handling complex occlusions or actions with subtle inter-class differences.

To address the above issues, we further propose the LSE. Building upon physical connectivity priors, the module first computes potential joint correlations via association energy. Subsequently, a dynamic gating mechanism is employed to adaptively strengthen discriminative local joint relationships, thereby capturing action-specific key spatial patterns. Finally, the physical mask, dynamic gating weights, and association energy are fused. To prevent the dynamic gate from excessively filtering useful information, we introduce an adaptively weighted residual connection, thereby generating locally enhanced spatial features, \( A_{\textit{l}} \), as formulated in Equation \ref{eq_Al}.
\begin{equation}
\label{eq_Al}
\begin{split}
E &= \bar{X}^{\top} \bar{X} \\
G &= \sigma(\textit{MLP}(\textit{Flatten}(\bar{X}))) \\
A_{\textit{l}} &= M \odot G \odot \textit{ReLU}(E) + \gamma \cdot M
\end{split}
\tag{6}
\end{equation}
where Flatten(\(\cdot\)) flattens the vector, MLP(\(\cdot\)) denotes a multi-layer perceptron, \( \sigma \) denotes the Sigmoid activation function, \( \odot \) denotes element-wise multiplication, and ReLU(\(\cdot\)) ensures non-negative association energy, \( \gamma \) is a learnable scalar parameter.

We employ weighted fusion between the global spatial feature topology matrix \( A_{\text{g}} \) and the local spatial feature topology matrix \( A_{\text{l}} \) to obtain the composite topology matrix \( A_s \). Subsequently, the feature transformation function converts the input skeleton feature \( X_{\text{s}} \in \mathbb{R}^{C_{\textit{in}} \times T \times N} \) into the senior representation bone feature \( X^i_{\text{s}} \in \mathbb{R}^{C_{\textit{out}} \times T \times N} \). Finally, all channel-wise output features are concatenated to obtain the final output \( Z_s \), as formulated in Equation \ref{eq_Zs}.
\begin{equation}
\label{eq_Zs}
\begin{split}
A_s &= \lambda_{\textit{g}} A_{\textit{g}} + \lambda_{\textit{l}} A_{\textit{l}} \\
Z_s &= \textit{Concat}\left(A_s^1 X_{\textit{s}}^1, A_s^2 X_{\textit{s}}^2, \dots, A_s^{C_{\textit{out}}} X_{\textit{s}}^{C_{\textit{out}}}\right)
\end{split}
\tag{7}
\end{equation}
where \( \lambda_{\text{g}} \) and \( \lambda_{\text{l}} \) are learnable fusion weights that dynamically balance the contributions of global dependencies and local structural features in the final spatial representation.

\subsection{Dual-path Hierarchical Temporal Module}
\begin{figure}[pos=htbp]
    \centering 
    \includegraphics[width=0.9\linewidth]{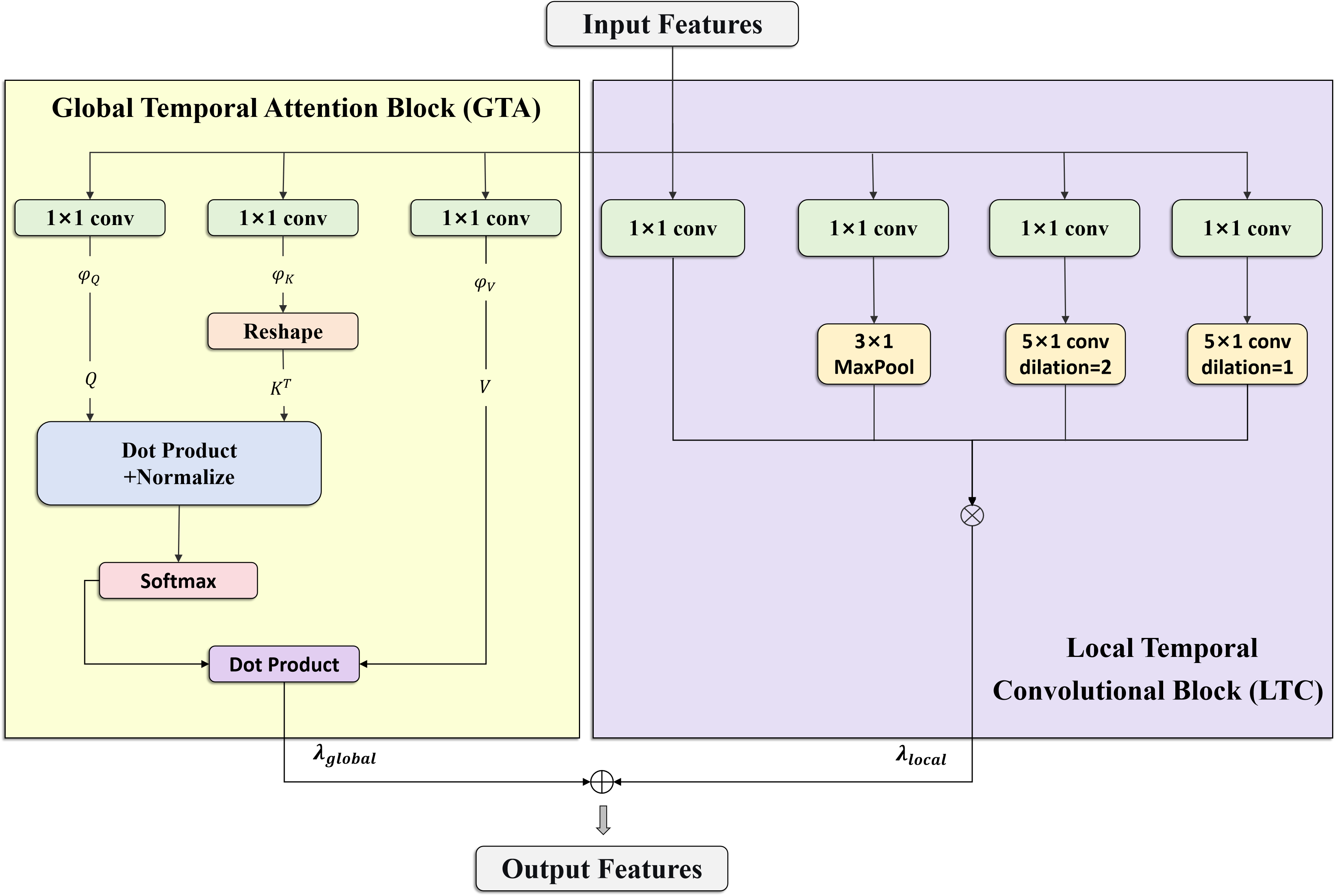}
    \caption{Overall architecture of the DHT. The yellow region on the left half corresponds to the GTA, and the purple region on the right half corresponds to the LTC.}
    \label{fig_DHT}
\end{figure}

To comprehensively model both long-range global dependencies and fine-grained local dynamics in skeleton actions, the DHT is proposed, as shown in Figure \ref{fig_DHT}. The component adopts a parallel dual-branch design: the Global Temporal Attention module (GTA) branch leverages the self-attention mechanism to flexibly capture long-range semantic dependencies across arbitrary frames, while the Local Temporal Convolutional module (LTC) branch efficiently extracts short-range temporal patterns by exploiting the inductive bias of convolution. The outputs of the two branches are combined through adaptive weighted fusion to form complementary temporal representations.

Traditional methods rely on global shared spatial topology and struggle to capture fine-grained temporal heterogeneity within each frame (\cite{18,22,27}). To address this issue, the GTA is proposed. Specifically, three linear transformations \( \varphi_Q(\cdot) \), \( \varphi_K(\cdot) \), and \( \varphi_V(\cdot) \) are first applied to the input skeleton feature \( X_{\textit{s}} \in \mathbb{R}^{C_{\textit{out}} \times T \times N} \) to obtain high-level feature \( \tilde{X} \in \mathbb{R}^{C_{\textit{in}} \times T \times N} \). Then, the temporal correlation between any two frames is computed point-wise to derive the global temporal attention network topology \( S_{\text{g}} \), as formulated in Equation \ref{eq_Sg}.
\begin{equation}
\label{eq_Sg}
S_{\textit{g}} = \textit{Softmax}\left( \frac{Q K^T}{\sqrt{C} \cdot N} \right) V = S V
\tag{8}
\end{equation}
where \( S \in \mathbb{R}^{T \times T} \) denotes the temporal attention map, and each element \( S_{ij} \) dynamically indicates the influence weight of the $i$-th frame exerts on the \(j\)-th frame.

The LTC is designed to model short-range temporal dependencies in skeleton sequences. For the input feature \( X' \in \mathbb{R}^{C \times T \times N} \), the module constructs four parallel temporal convolutional branches, formulated as Equation \ref{eq_S}.
\begin{equation}
\label{eq_S}
\begin{split}
S_1 &= \textit{Conv1D}(K=5, d=1)\bigl(\textit{Conv}1\times1(X')\bigr) \\
S_2 &= \textit{DilatedConv1D}(K=5, d=2)\bigl(\textit{Conv}1\times1(X')\bigr) \\
S_3 &= \textit{MaxPool}(K=3, s=1)\bigl(\textit{Conv}1\times1(X')\bigr) \\
S_4 &= \textit{Conv}1\times1(X')
\end{split}
\tag{9}
\end{equation}
where \(K\) is the kernel size of the convolution operation, and \(d\) is the dilation rate of the convolution operation, \(s\) is the stride in the convolution operation. \( S_1 \) is the standard convolutional branch used to capture local motion of adjacent frames; \( S_2 \) is the dilated convolutional branch for expanding the receptive field during periodic mode; \( S_3 \) is the pooling branch that enhances robustness against sudden motion variations by maximum pooling; and \( S_4 \) is the \( 1 \times 1 \) convolutional branch for channel-wise information adjustment and alignment. The outputs of the four branches are then concatenated to obtain the local multi-scale temporal \( S_{\text{l}} = \textit{Concat}(S_1, S_2, S_3, S_4) \).

Finally, we introduce learnable fusion weights \( \lambda_{\textit{g}} \) and \( \lambda_{\textit{l}} \) to dynamically balance the occupancy of global and local temporal features in the fusion process, thereby obtaining the final temporal representation of DHT, as formulated in Equation \ref{eq_St}.
\begin{equation}
\label{eq_St}
S_{\textit{T}} = \lambda_{\textit{g}} X_{\textit{g}} + \lambda_{\textit{l}} X_{\textit{l}}
\tag{10}
\end{equation}
\subsection{Skeleton-language Sequential Fusion Module}
\begin{figure}[pos=htbp]
    \centering 
    \includegraphics[width=0.8\linewidth]{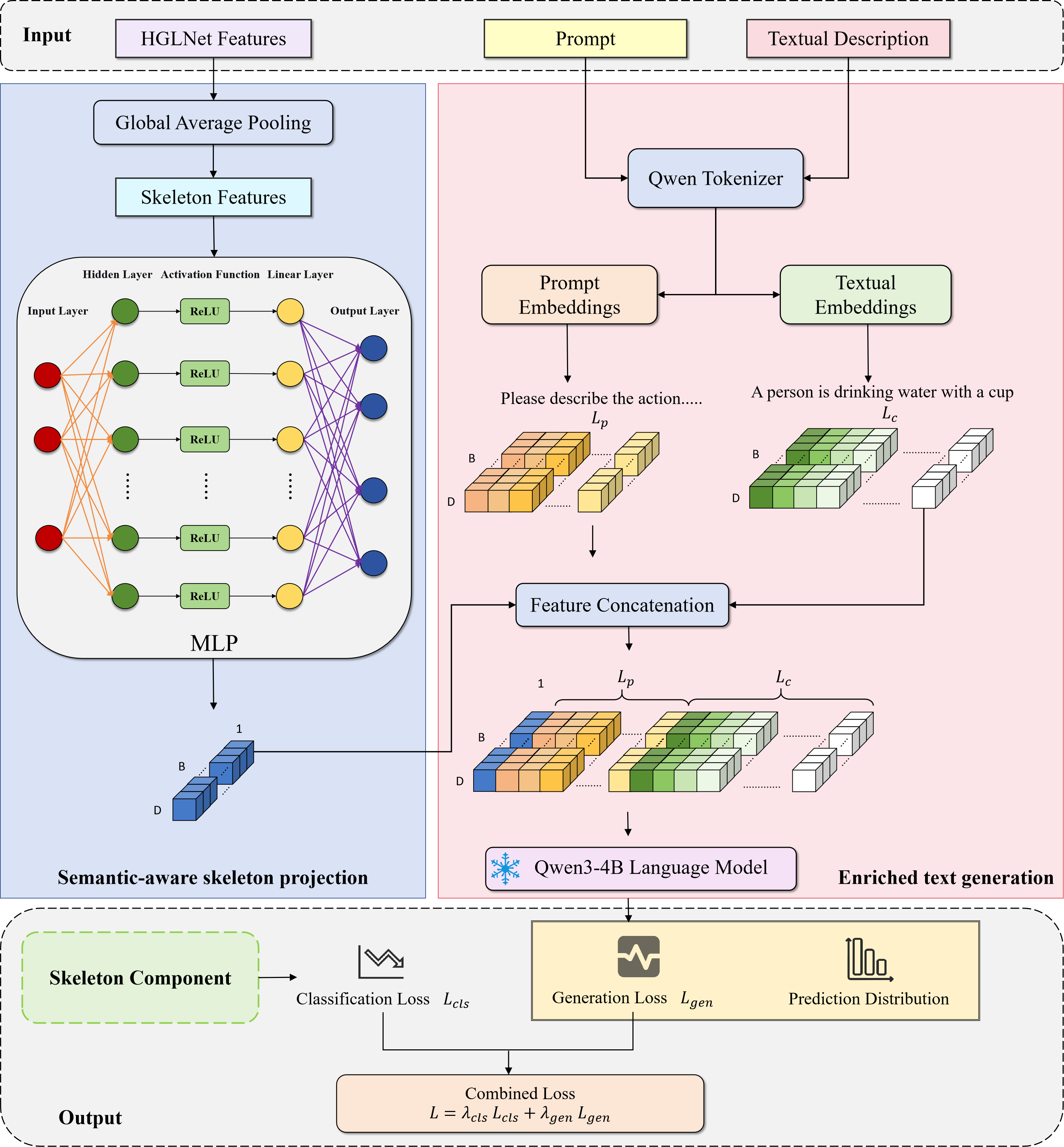}
    \caption{The overall architecture of SSF.}
    \label{fig_SSF}
\end{figure}

Traditional skeleton-based action recognition mainly relies on spatio-temporal features, lacking the ability to model action semantics, which leads to limited performance in cross-category generalization and semantic consistency (\cite{20,21,26}). To enhance the semantic expressiveness of skeleton action features, we introduce SSF to combine the skeletal feature and semantic feature. As illustrated in Figure \ref{fig_SSF}, the model projects skeleton features into the semantic embedding space of the Qwen 3 large language model, enabling the model to simultaneously learn the spatio-temporal representations of actions and their corresponding linguistic semantic associations during training, thereby enhancing semantic understanding.

To achieve deep understanding of skeleton semantics, we first build the skeleton semantic understanding block. We extract spatio-temporal features from the skeleton sequence using HGLNet, mapping the raw coordinate space to a high-dimensional semantic space. To obtain a compact global description, global average pooling is performed along both temporal dimension \(T\) and spatial dimension \(N\), yielding the sample-level skeleton semantic vector \( f_s \in \mathbb{R}^{B \times D} \), as formulated in Equation \ref{eq_fs}.
\[
\label{eq_fs}
f_s = \textit{GAP} \left( \textit{HGLNet}(X_{\textit{t}}) ;\tau, \nu \right)
\tag{11}
\]
where \( X_{\text{t}} \) denotes the input skeleton sequence, and \( \textit{GlobalAvgPool}(\cdot) \) represents the global average pooling over the temporal dimension $\tau$ and spatial dimension $\nu$.

We adopt a double-layer MLP projection unit to map the spatiotemporal features extracted by the skeleton encoder to the semantic space aligned with the language model. Specifically, as shown in Equation \ref{eq_[SK]}, the first linear transformation with parameters \( W_1 \in \mathbb{R}^{D \times D}, b_1 \in \mathbb{R}^D \) performs channel mixing on the skeleton features, followed by ReLU activation to introduce non-linearity. Then, the second linear transformation with \( W_2 \in \mathbb{R}^{D \times D^*}, b_2 \in \mathbb{R}^{D^*} \) with higher dimension $D^*$ is applied, and LayerNorm ensures numerical stability while maintaining the consistent feature distribution with the pretrained language model space, yielding the visual token \( [SK] \in \mathbb{R}^{B \times 1 \times D^*} \) aligned with the text embedding space.
\begin{equation}
\label{eq_[SK]}
\begin{aligned}
&h = \textit{ReLU}(W_1 f_s + b_1) \\
&[SK] = f_{\textit{proj}} = \textit{LayerNorm}(W_2 h + b_2)
\end{aligned}
\tag{12}
\end{equation}

In the subsequent conditional text generation block, to generate natural language descriptions of the actions, the visual token \( [SK] \) obtained from the previous step is concatenated with a learnable text prompt embedding \( E_{\text{p}} \) and the textual description embedding \( E_{\text{t}} \), forming a unified multimodal input sequence, as formulated in Equation \ref{eq_X}.
\[
\label{eq_X}
X = \textit{Concat}([SK], E_{\textit{p}}, E_{\textit{t}})
\tag{13}
\]
where \( [SK] \in \mathbb{R}^{B \times 1 \times D^*} \) is the visual token derived from projecting the skeleton spatio-temporal features, \( E_{\text{prompt}} \in \mathbb{R}^{B \times L_p \times D^*} \) is a learnable fixed prompt embedding used to guide the language model in generating action descriptions, and \( E_{\text{t}} \in \mathbb{R}^{B \times L_t \times D^*} \) is the embedding of the action text caption. After concatenation, the multimodal input \( X \in \mathbb{R}^{B \times (1 + L_p + L_t) \times D^*} \) is fed into the language model to achieve the alignment and fusion of visual features with linguistic context in the unified semantic space.

Qwen3-4B is a pre-trained language model based on the autoregressive mechanism, whose core task is to generate coherent and contextually consistent natural language texts according to given input conditions. In autoregressive models, the text generation process is carried out incrementally, where the output at each step depends on the previously generated content, thereby ensuring that the final generated text maintains semantic and contextual consistency and coherence. First, the concatenated multimodal input sequence $X$ is fed into the Qwen3-4B model for processing. During this process, the input is first transformed through an embedding layer, which maps text prompts and action descriptions into continuous vector representations. In particular, the visual token $[\text{SK}]$ is embedded into the model as a special token, carrying the spatiotemporal information of skeleton data. This token plays a crucial role in multimodal input: it provides the model with key visual features, enabling the effective fusion of skeleton information and text information, thus enhancing the model's capability to generate action descriptions. Second, Qwen3-4B generates action descriptions using the autoregressive mechanism. The model relies on a multi-layer Transformer architecture and generates a new word at each time step in an incremental manner. Specifically, the model generates the initial part of the description based on the visual token $[\text{SK}]$ and the text prompt embedding $E_p$. Subsequently, the model continues to generate subsequent texts by utilizing the already generated text segment and the action description embedding $E_t$. Each generation stage depends on the output of the preceding text; this autoregressive mechanism ensures the coherence and semantic consistency of the generated text. Finally, during the training process, the objective of Qwen3-4B is to maximize the conditional probability of generating the correct action description given the input condition. Specifically, the model calculates the discrepancy between the generated text and the ground-truth action description, and optimizes the model parameters by minimizing the loss. This process is implemented via the backpropagation algorithm, where parameters are adjusted incrementally to improve the accuracy and consistency of the generated text. Through this training process, Qwen3-4B learns to effectively generate natural language descriptions that match visual features, and can organically integrate the spatiotemporal features of skeleton data with the contextual information of actions, thereby producing more natural and semantically accurate descriptions.

To enable the overall model to learn from both generative and classification tasks, we propose a joint optimization strategy in the collaborative optimization block. The generative loss \( L_{\text{gen}} \) measures the difference between the generated text and the true action description. The classification loss \( L_{\text{cls}} \) ensures accurate action recognition through the backbone features. The final total loss is the weighted sum of both losses. As given in Equation \ref{eq_total}.
\begin{equation}
\label{eq_total}
\begin{split}
L_{\textit{gen}} &= -\frac{1}{N} \sum_{i=1}^N \sum_{t=1}^T \mathbf{1}_{y_{i,t} \neq -100} \log P_{\textit{qwen}}(y_{i,t} \mid X_{i,<t}) \\
L_{\textit{cls}} &= \textit{CrossEntropy}(\textit{Classifier}(f_s), y_{\textit{true}}) \\
L_{\textit{total}} &= \lambda_{\textit{gen}} L_{\textit{gen}} + \lambda_{\textit{cls}} L_{\textit{cls}}
\end{split}
\tag{14}
\end{equation}
where \( P_{\textit{qwen}} \) denotes the probability of the model predicting a single token under the given input condition \( X_{i,<t} \). The Classifier(\( f_s \)) represents the action classification network based on the skeleton feature \( f_s \), and \( y_{\textit{true}} \) is the true action category label. \( \lambda_{\textit{gen}} \) and \( \lambda_{\textit{cls}} \) are hyperparameters that balance the importance of the generative and recognition tasks.

\section{Experiment}
\label{sec_experiment}
\subsection{Experimental Setup}
\subsubsection{Datasets}
In this context, we employ three action recognition datasets, i.e., NTU RGB+D (\cite{43}), NTU RGB+D 120 (\cite{44}), Northwestern-UCLA (\cite{45}), to evaluate our proposed method, which is summarized as follows:
\begin{itemize}
    \item NTU RGB+D (\cite{43}) is a core benchmark dataset in 3D action recognition, containing 56,880 samples across 60 action classes performed by 40 subjects. The data are synchronously captured by three Kinect v2 cameras from different viewpoints and represented as 3D coordinates of 25 joints per frame. The dataset provides two standard evaluation protocols: Cross-Subject (X-Sub), which evenly splits subjects into training and test sets; and Cross-View (X-View), which uses data from cameras 2 and 3 for training and data from camera 1 for testing, to evaluate the model’s generalization to novel viewpoints.
    \item NTU RGB+D 120 (\cite{44}) is an extended version of NTU RGB+D and currently constitutes one of the largest skeleton-based 3D action recognition benchmarks. It comprises 120 action classes and over 113,000 samples, collected from 106 subjects under 32 different capture setups using three Kinect cameras. Two standard evaluation protocols are adopted: Cross-Subject (X-Sub), which evenly divides subjects for training and testing; and Cross-Setup (X-Setup), which splits the dataset according to the parity of setup IDs to assess model robustness to unseen scenarios.
    \item Northwestern-UCLA (\cite{45}) is an important benchmark for multi-view action recognition, consisting of 1,494 samples across 10 human action categories performed by 10 subjects and synchronously recorded by three Kinect cameras from different viewpoints. The dataset follows the standard evaluation protocol: samples captured by the first two cameras are used for training, while those from the third camera serve as the test set, thereby examining the model’s cross-view generalization capability.
\end{itemize}

\subsubsection{Evaluation Metric}
In this study, we follow the mainstream evaluation standard in the field of skeleton-based action recognition and adopt Top-1 Accuracy as the core performance metric. Top-1 Accuracy directly reflects the proportion of test samples for which the class predicted with the highest confidence matches the ground-truth label. It serves as an intuitive yet rigorous measure of a classification model's discriminative ability. The formula is defined as:
\[
\text{Top-1 Accuracy} = \frac{1}{N} \sum_{i=1}^{N} \mathbb{I}(\arg\max(\mathbf{p}_i) = y_i) \times 100\%
\tag{15}
\]
where $N$ denotes the total number of test samples, $\mathbf{p}_i$ is the predicted probability vector for the $i$-th sample, $y_i$ is its true class label, and $\mathbb{I}(\cdot)$ is the indicator function (equal to 1 if the prediction is correct, and 0 otherwise).

\subsection{Implementation Details}
\begin{figure}[pos=htbp]
    \centering
    \includegraphics[width=0.8\linewidth]{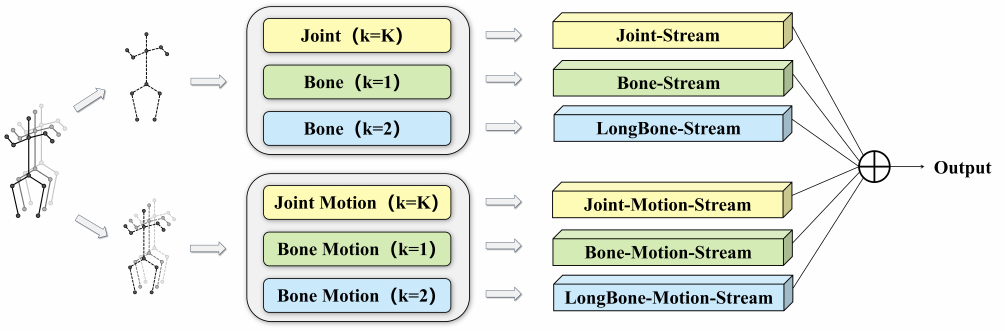}
    \caption{The overall architecture of 6-stream ensemble.}
    \label{fig_stream}
\end{figure}

All experiments in this paper are conducted on NVIDIA GeForce RTX Pro 6000 GPUs using the PyTorch platform. The model adopts a multi-stream ensemble architecture consisting of six data streams: joint stream, first-order bone stream, second-order bone stream, and their corresponding three motion streams, as shown in Figure \ref{fig_stream}. The data preprocessing pipeline strictly follows the standard protocol of CTR-GCN (\cite{29}) to ensure fair comparison. For different dataset characteristics, the training and testing batch size for NTU RGB+D, NTU RGB+D 120, and Northwestern-UCLA is uniformly set to 64, with a total of 85 training epochs to ensure sufficient convergence and optimal performance. Model training jointly optimizes the cross-entropy classification loss and the generative loss from the SSF module via a multi-task learning framework, where the weights for generative loss and classification loss are \( \lambda_{\text{gen}} \) and \( \lambda_{\text{cls}} \), respectively. The AdamW optimizer is employed with an initial learning rate of \( 1 \times 10^{-4} \) and a weight decay coefficient of 0.01 to enhance model generalization.

\subsection{Comparison With the State-of-the-Art}

According to the results in Table \ref{tab:NTU-RGB+D}, Table \ref{tab:nturgbd120}, and Table \ref{tab:northwestern-ucla}, HocSLM achieves high accuracy across evaluation settings and significantly outperforms other methods. 
\begin{table}
  \caption{Comparison of different models on the NTU-RGB+D dataset.}
  \label{tab:NTU-RGB+D}
  \begin{tabular}{lcc}
    \toprule
    Method & X-Sub Acc (\%) & X-View Acc (\%) \\
    \midrule
    ST-GCN \cite{18} & 81.5 & 88.3 \\
    AS-GCN \cite{46} & 86.8 & 94.2 \\
    2s-AGCN \cite{27} & 88.5 & 95.1 \\
    SGN \cite{47} & 89.0 & 94.5 \\
    AGC-LSTM \cite{48} & 89.2 & 95.0 \\
    DGNN \cite{49} & 89.9 & 96.1 \\
    ST-TR-agcn \cite{50} & 90.3 & 96.3 \\
    Shift-GCN \cite{22} & 90.7 & 96.5 \\
    DC-GCN + ADG \cite{51} & 90.8 & 96.6 \\
    Dynamic-GCN \cite{28} & 91.5 & 96.0 \\
    MS-G3D \cite{20} & 91.5 & 96.2 \\
    MST-GCN \cite{21} & 91.5 & 96.6 \\
    CTR-GCN \cite{29} & 92.4 & 96.8\\
    GAP \cite{52} & 92.9 & 97.0 \\
    Info-GCN \cite{26} & 93.0 & 97.1 \\
    LA-GCN \cite{30} & \underline{93.5}& \textbf{97.2} \\
    \textbf{HocSLM (Ours)} & \textbf{93.6} & \underline{97.1}\\
    \bottomrule
  \end{tabular}
\end{table}
\begin{table}[ht]
  \caption{Comparison of different models on the NTU-RGB+D120 dataset.}
  \label{tab:nturgbd120}
  \begin{tabular}{lcc}
    \toprule
    Method & X-Sub Acc (\%) & X-View Acc (\%) \\
    \midrule
    ST-GCN \cite{18} & 70.7 & 73.2 \\
    2s-AGCN \cite{27} & 82.5 & 84.2 \\
    SGN \cite{47} & 79.2 & 81.5 \\
    ST-TR-agcn \cite{50} & 85.1 & 87.1 \\
    Shift-GCN \cite{22} & 85.9& 87.6 \\
    DC-GCN + ADG \cite{51} & 86.5 & 88.1\\
    MS-G3D \cite{20} & 86.9 & 88.4 \\
    Dynamic-GCN \cite{28} & 87.3 & 88.6 \\
    MST-GCN \cite{21} & 87.5 & 88.8 \\
    CTR-GCN \cite{29} & 88.9 & 90.6 \\
    Info-GCN \cite{26} & 89.4 & 90.7 \\
    GAP \cite{52} & 89.9 & 91.1 \\
    LA-GCN \cite{30} & \underline{90.7}& \textbf{91.8}\\
    \textbf{HocSLM (Ours)} & \textbf{90.9} & \textbf{91.8} \\
    \bottomrule
  \end{tabular}
\end{table}

\begin{table}[ht]
  \caption{Comparison of different models on the Northwestern-UCLA dataset.}
  \label{tab:northwestern-ucla}
  \begin{tabular}{lc}
    \toprule
    Method & Acc (\%) \\
    \midrule
    Shift-GCN \cite{22} & 94.6 \\
    GSTLN \cite{53} & 94.8 \\
    CTR-GCN \cite{29} & 96.5 \\
    Info-GCN \cite{26} & 96.6 \\
    MSS-GCN \cite{54} & 96.6 \\
    SAN-GCN \cite{55} & 96.8 \\
    BlockGCN \cite{56} & 96.9 \\
    Info-GCN \cite{26} & 97.0 \\
    UGSTM \cite{57} & \underline{97.2}\\
    \textbf{HocSLM (Ours)} & \textbf{97.4} \\
    \bottomrule
  \end{tabular}
\end{table}

On the NTU RGB+D dataset, HocSLM surpasses CTR-GCN (\cite{29}) by 1.2\% on the X-Sub benchmark and by 0.3\% on the X-View benchmark. This achievement is attributed to HocSLM's dynamic global and local spatio-temporal modeling through HGLNet, which effectively captures long-range joint dependencies and local details. In comparison, HocSLM outperforms LA-GCN (\cite{30}) (93.5\%) by 0.1\% on the X-Sub benchmark, while slightly falling short on the X-View benchmark, with a difference of only 0.1\%. Overall, the performance of both models is comparable, each with its advantages. The slight performance difference stems from the fundamental differences in the ways they integrate LLMs and interact with knowledge. LA-GCN uses BERT as a static source of prior knowledge, extracting textual features to construct fixed topologies (GPR and CPR). It does not generate text and cannot adapt to the dynamic changes in skeleton sequences. Additionally, the language prior is unidirectional and injected once at the start, without any further interaction with the skeleton features during training. This design allows LA-GCN to effectively integrate linguistic prior knowledge with skeleton topology modeling, greatly enhancing the semantic relationships between skeleton nodes and significantly improving the accuracy of topological modeling. This is especially effective in distinguishing actions with similar node relationships, which is why it slightly outperforms HocSLM on the X-View benchmark. However, due to the static nature of the prior, it lacks dynamic spatio-temporal modeling capabilities, limiting its ability to handle complex spatio-temporal dependencies in action sequences.

In contrast, HocSLM uses Qwen3-VL/ Qwen3-4B as dynamic semantic interaction modules. Qwen3-VL generates matching action descriptions, which are then dynamically aligned with skeleton spatio-temporal features through the SSF module, enabling real-time cross-modal interaction. This design allows for bidirectional, end-to-end collaborative optimization between the LLM and skeleton features. After the skeleton features are mapped to the LLM's semantic space and fused with the text, the model simultaneously optimizes both classification and generation losses, with each improving the other. This approach gives HocSLM a significant advantage in spatio-temporal modeling, enabling it to handle complex spatio-temporal dependencies more effectively, and thereby achieving a slight improvement over LA-GCN on the X-Sub benchmark.

On the NTU RGB+D 120 dataset, HocSLM achieves an accuracy of \textbf{90.9\%} on the X-Sub benchmark and \textbf{91.8\%} on the X-Set benchmark, demonstrating its stability and semantic exploration ability on large-scale datasets. In contrast, while LA-GCN excels in recognizing similar action categories, its performance is limited by the semantic empowerment logic of the LLM. LA-GCN utilizes LLM only to enhance the semantic relationships between skeleton nodes, addressing the topology modeling issue, but its empowerment is confined to the spatial relationships of nodes and does not account for the spatio-temporal dynamic semantics of actions. Additionally, relying on a static topological structure, it fails to capture the fine details of spatio-temporal dependencies, which restricts its performance on large-scale, complex datasets. In comparison, HocSLM leverages LLM to achieve full-dimensional semantic alignment and task generalization of skeleton actions. It maps the spatio-temporal dynamic features of skeletons and natural language semantics to a unified semantic space, allowing the model to learn both geometric and motion features of the skeleton, understand high-level semantics of actions, and capture spatio-temporal dependencies and fine-grained semantic information of long-sequence actions. By combining the multi-scale spatio-temporal modeling of HGLNet and the cross-modal semantic alignment of the SSF module, HocSLM significantly enhances the semantic understanding of complex actions, leading to superior performance in complex scenarios.

On the Northwestern-UCLA dataset, HocSLM achieves an accuracy of \textbf{97.4\%}, surpassing UGSTM (97.2\%). While UGSTM enhances performance through uncertainty-guided spatio-temporal modeling, HocSLM leverages the integration of LLM with dynamic spatio-temporal modeling, along with the cross-modal semantic alignment via the SSF module, significantly improving semantic understanding and the capture of spatio-temporal dependencies. This leads to superior performance, especially in complex action and long-sequence data recognition tasks.

Overall, HocSLM combines the spatio-temporal modeling of HGLNet with the semantic alignment of SSF, achieving a breakthrough in traditional LLM integration by dynamically merging the LLM. This design enhances the model's ability to capture spatial and temporal dependencies while significantly improving semantic understanding and task generalization. This is the key to its superior performance across the NTU RGB+D, NTU RGB+D 120, and Northwestern-UCLA datasets, demonstrating that HocSLM not only outperforms other methods in terms of accuracy but also exhibits strong generalization capabilities, making it highly suitable for a wide range of action recognition tasks.

\subsection{Ablation Study}
\subsubsection{Contribution of Core Components}
To validate the effectiveness of the three proposed core modules, this paper uses CTR-GCN as the baseline and conducts module-wise ablation studies on the NTU RGB+D Cross-Subject (X-Sub) benchmark using the joint-only (k=K) setting. The results are reported in Table \ref{tab:ablation_core}.
\begin{table}[htbp]
\centering
\caption{Fine-grained Ablation on Different Core Components.}
\label{tab:ablation_core}
\begin{tabular}{cccccc}
\toprule
Configuration & CTS & DHT & SSF & Top-1 (\%) & Performance\\
(Baseline + Components) & & & & & enhancement(\%)\\
\midrule
Baseline (CTR-GCN) & - & - & - & 89.96 & - \\
Baseline + CTS & $\checkmark$ & - & - & 90.31 & 0.35 \\
Baseline + CTS + DHT & $\checkmark$ & $\checkmark$ & - & 90.63 & 0.67\\
\textbf{HocSLM}& $\checkmark$ & $\checkmark$ & $\checkmark$ & \textbf{90.84}& \textbf{0.88}\\
\bottomrule
\end{tabular}
\end{table}

The experimental results demonstrate the effectiveness of the three core modules of the HocSLM framework in enhancing the action recognition performance. First, introducing the CTS module significantly improves spatial representation by dynamically modeling joint associations, yielding a 0.35\% increase in accuracy. This enhancement allows the model to better capture the spatial relationships between joints, which is crucial for understanding complex actions. Next, the DHT module is incorporated, which introduces a dual-path approach for temporal modeling. This module further improves the model’s ability to capture long-range temporal dependencies and fine-grained dynamics within each action sequence, providing an additional 0.32\% improvement. Finally, the SSF module is added to introduce cross-modal semantic supervision, which significantly boosts the model’s ability to discriminate fine-grained actions. This module aligns skeletal spatio-temporal features with textual descriptions, providing a 0.21\% improvement. Together, these components contribute a total of 0.88\% enhancement, achieving a final Top-1 accuracy of \textbf{90.84\%}. These results highlight the complementary advantages of dynamic spatial topology modeling, hierarchical temporal modeling, and cross-modal semantic enhancement, underscoring the overall effectiveness of the HocSLM framework in action recognition tasks. 

\subsubsection{Contribution Analysis of HGLNet Components}
To verify the effectiveness and complementary roles of the global and local modeling components within HGLNet, this paper adopts CTR-GCN as the baseline and conducts a series of fine-grained ablation studies on the NTU RGB+D 60 Cross-Subject benchmark using the joint-only (k=K) setting. HGLNet consists of two main parts: (1) the CTS, comprising GSM and LSE; (2) the DHT, comprising GTA and LTC. The contribution of each component is reported in Table \ref{tab:ablation_hstnet}.
\begin{table}[htbp]
\centering
\caption{Fine-grained ablation on HGLNet components.}
\label{tab:ablation_hstnet}
\begin{tabular}{cccc}
\toprule
Configuration & CTS Module& DHT Module& Top-1 (\%) \\
\midrule
Baseline (CTR-GCN) & - & - & 89.96 \\
Only Local & Only LSE & Only LTC & 89.71 \\
w/o GSM & Only LSE & LTC + GTA & 90.20 \\
w/o GTA & LSE + GSM & Only LTC & 90.37 \\
\textbf{HGLNet}& \textbf{LSE + GSM}& \textbf{LTC + GTA}& \textbf{90.63}\\
\bottomrule
\end{tabular}
\end{table}

The experimental results show that removing any global or local component leads to performance degradation. In particular, removing GSM and GTA causes accuracy drops of 0.43\% and 0.26\%, respectively, indicating their significant impact. When only the local components are used, the model achieves an accuracy of 89.71\%, which is 0.92\% lower than the full HGLNet model. This highlights the important role of the global components in improving the model's performance. In contrast, the complete HGLNet design, with its parallel global-local collaborative approach, ultimately achieves \textbf{90.63\%} accuracy. This demonstrates the strong complementarity between the global and local branches in dynamic spatial topology modeling and long-range dependency capture. The GSM and GTA components are especially crucial, as they enable the model to capture complex long-range dependencies and fine-grained temporal dynamics, which are essential for high-accuracy action recognition. Overall, the ablation study shows that both global and local components are indispensable for the optimal performance of the HocSLM framework, emphasizing the importance of dynamic spatio-temporal feature modeling in action recognition.

\subsubsection{Training Paradigms and Effectiveness Conditions of SSF}

To validate the role of SSF in enhancing skeleton-based action recognition, we designed four different training strategies and conducted ablation experiments focusing on freeze strategies and loss combinations. All experiments were conducted on the NTU RGB+D 60 X-Sub (Joint modality) dataset, using a model architecture consisting of SKeleton Encoder (SKE) + Semantic Projection Layer (MLP) + LLM. By adjusting the freeze settings of SKE, MLP, and LLM, and combining language generation loss and classification loss, we analyzed the impact of semantic priors on action features. HGLNet (89.96\%) was used as the baseline, and the experimental results are shown in Table \ref{tab:ablation_training}.

\begin{table}[htbp]
\centering
\caption{Ablation on training paradigms and effectiveness conditions of SSF.}
\label{tab:ablation_training}
\begin{tabular}{cccccc}
\toprule
Strategy & Skeleton Encoder & MLP & LLM & Training Loss & Top-1 (\%) \\
\midrule
T0 (Baseline) & Trained (HGLNet) & -- & -- & $\mathcal{L}_{\text{cls}}$ & 90.63 \\
T1 & Frozen $\to$ Unfrozen & Trained & Frozen & Stage1: $\mathcal{L}_{\text{gen}}$ only & 89.08 \\
    &                     &        &        & Stage2: $\mathcal{L}_{\text{cls}}$ only & \\
T2 & Trained & Trained & Frozen & $\mathcal{L}_{\text{gen}} \rightarrow \mathcal{L}_{\text{cls}}$ & 90.73 \\
\textbf{T3 (Ours)}& \textbf{Trained}& \textbf{Trained}& \textbf{Frozen}& \textbf{$\mathcal{L}_{\text{gen}} + \mathcal{L}_{\text{cls}}$}& \textbf{90.84}\\
T4 & Trained from scratch & Trained & Frozen & $\mathcal{L}_{\text{gen}} + \mathcal{L}_{\text{cls}}$ & 68.23 \\
\bottomrule
\end{tabular}
\end{table}
\begin{itemize}

    \item \textbf{Strategy 1 (T1)}: Freeze SKE and LLM, train MLP (language generation loss), then unfreeze SKE+MLP for classification training. Initially, SKE and LLM were frozen, and only MLP was trained for language generation. The language modeling process was successful, and MLP learned the mapping from skeleton features to language space. However, the large output dimensionality of MLP (2560 dimensions for the Qwen3-4B embedding vector) led to overfitting when SKE+MLP was unfrozen for classification training, causing the model to memorize noise rather than extracting useful features for action recognition. As a result, the model achieved \textbf{89.08\%} accuracy on the test set, which is lower than the baseline. 
\end{itemize}
\begin{itemize}
  \item \textbf{Strategy 2 (T2)}: Freeze LLM, train SKE+MLP (language generation loss), then use the trained SKE for classification training. In this strategy, SKE is fine-tuned for the language generation task, but due to the lack of explicit classification supervision, the adjustments made to SKE mainly serve the language generation goal, rather than being aligned with action category discrimination. As a result, the model achieved \textbf{90.73\%} accuracy, showing a slight improvement.
  \item \textbf{Strategy 3 (T3 - Ours)}: Freeze LLM, train SKE+MLP (language generation loss + classification loss). When both losses are involved in training, SKE and MLP are optimized under dual supervision. The classification loss helps enhance features useful for category discrimination, while the generation loss injects language semantic structure. The two losses complement each other, significantly improving the model's generalization performance. This strategy is the only one among the four that consistently improves test set accuracy, achieving a Top-1 accuracy of \textbf{90.84\%}, demonstrating that cross-modal semantic alignment must be optimized alongside task-specific classification supervision to achieve optimal results. 
  \item \textbf{Strategy 4 (T4)}: Freeze LLM, use a randomly initialized SKE, train SKE+MLP (generation loss + classification loss). This strategy explores whether cross-modal semantic supervision can work without pre-training the skeleton encoder. The results show that with random initialization, SKE struggles to learn robust spatio-temporal representations when both losses are used, achieving an accuracy of \textbf{68.23\%}. This indicates that pre-trained action representations in the skeleton encoder are necessary for semantic supervision to be effective.
\end{itemize}

\subsubsection{Multi-stream Ensemble}
This subsection ensembles skeleton representations generated from six different input modalities (Joint, Bone, Joint Motion, Bone Motion, Joint 2nd-order, and Vel 2nd-order), as shown in Table \ref{tab:ablation_multistream}. On the NTU RGB+D 60 X-Sub benchmark, the six-stream ensemble achieves \textbf{93.61\%} accuracy, outperforming the conventional four-stream ensembles by 0.29\% and the strongest single stream (Bone) by 2.10\%. These results fully demonstrate the complementarity among multi-modal information and the effectiveness of the ensemble strategy.
\begin{table}[htbp]
\centering
\caption{Ablation on multi-stream fusion strategies.}
\label{tab:ablation_multistream}
\begin{tabular}{clc}
\toprule
Streams & Method & Top-1 (\%) \\
\midrule
\multirow{6}{*}{1s} 
& Joint only & 90.84 \\
& Bone only & 91.51 \\
& Joint Motion only & 89.15 \\
& Bone Motion only & 88.72 \\
& Joint 2nd-order only & 91.42 \\
& Vel 2nd-order only & 88.87 \\
\midrule
2s & Joint + Bone & 92.64 \\
\midrule
4s & Joint + Bone + Joint Motion + Bone Motion & 93.32 \\
\midrule
\textbf{6s}& \textbf{Joint + Bone + J-Motion + B-Motion + Joint 2nd-order + Vel 2nd-order}& \textbf{93.61}\\
\bottomrule
\end{tabular}
\end{table}
\subsection{Visualization}
\begin{figure}[pos=htbp]
    \centering
    \includegraphics[width=1.0\linewidth]{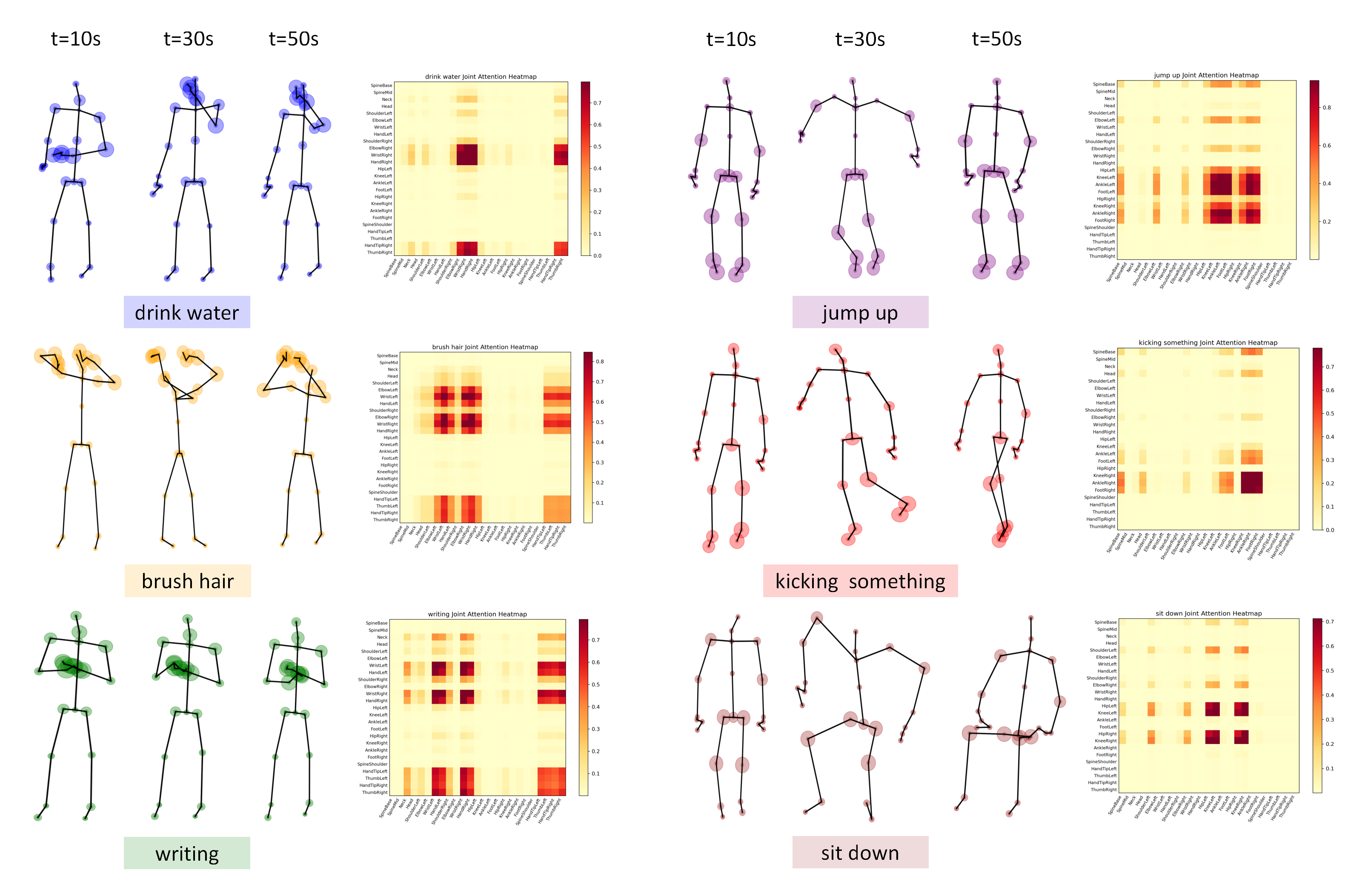}
    \caption{Visualization of skeleton diagrams and joint-Joint attention matrices for six typical actions.}
    \label{fig_skeleton}
\end{figure}

Figure \ref{fig_skeleton} visualizes the spatio-temporal attention distribution for six representative actions (drink water, brush hair, writing, jump up, kicking something, sit down) at three key frames sampled every 20 frames. Attention intensity is represented by the radius of circles centered on the joints, where the size of the circle indicates the impact of the corresponding joint on overall action recognition. The heatmaps illustrate the dynamically learned inter-joint interaction weights within the current frame, effectively capturing the relationships between joints. The experimental results demonstrate that the complete CTS module can adaptively generate long-range topological structures relevant to the action, precisely focusing on key joints, particularly those crucial for action recognition. In contrast, the pure local variant is limited to modeling within physical neighborhoods, failing to capture long-distance inter-joint interactions, while the pure global variant shows attention dispersion, leading to ambiguous recognition of joint relationships. These results fully validate the advantages of the global-local collaborative mechanism in spatial modeling, ensuring the model can efficiently focus on the most relevant spatio-temporal features, thereby significantly improving the model's action recognition capability. 

\begin{figure}[pos=htbp]
    \centering
    \includegraphics[width=0.85\linewidth]{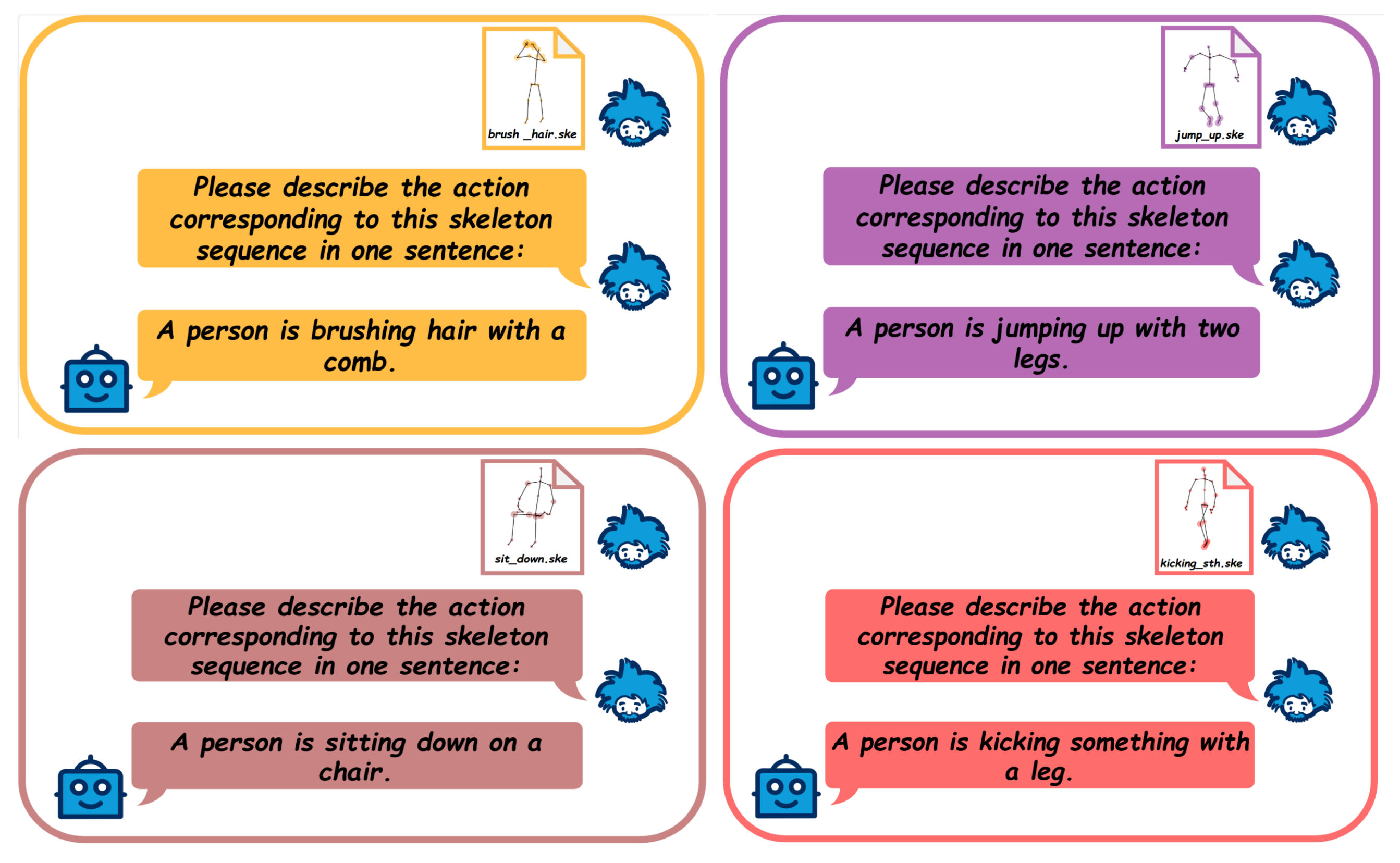}
    \caption{The generated text examples by Qwen3-4B based HocSLM.}
    \label{fig_Qwen3}
\end{figure}

To further validate the effectiveness of the alignment training strategy, a visualization analysis is conducted, and the results are shown in Figure \ref{fig_Qwen3}. 
The experiment constructs the training framework based on the large language model Qwen3-4B, with the backbone parameters of the language model fixed and only the skeleton feature extraction network and MLP projection layer being jointly trained. During the optimization process, an LLM-based loss function is employed to guide the model optimization, minimizing the semantic discrepancy between the generated action description and the ground truth labels. The visualization results demonstrate that, during alignment training, the model successfully maps dynamic skeleton features to the semantic representation space of the language model, achieving precise alignment between the visual skeleton sequence and the natural language action description. For example, when the model is provided with a skeleton sequence of brushing hair, it generates the text description ``A person is brushing hair with a comb'', which intuitively confirms that the proposed method can effectively establish the alignment between visual skeleton features and language semantics, enabling the language model to accurately understand the skeleton action input and generate highly matching textual descriptions. Furthermore, the alignment strategy during training enhances the model's cross-modal learning ability, making the generated text descriptions more accurate and semantically consistent.

\section{Conclusion}
\label{sec_conclusion}
In this study, we proposed a novel large action model (HocSLM) to address the limitations of existing GCN-based methods for human action recognition. First, we designed a hierarchical global-local network that combines a composite-topology spatial module and a dual-path hierarchical temporal module. This approach enabled dynamic, collaborative modeling at both global and local scales, significantly enhancing the model's ability to represent complex spatio-temporal relationships while preserving prior knowledge of human physical structure. Secondly, we introduced a skeleton-language sequential fusion module, leveraging large language models and jointly optimizing both generation and classification losses. This module ensured precise alignment between skeletal spatio-temporal features and action descriptions within a unified semantic space, significantly improving the model's semantic discrimination and cross-modal understanding capabilities. Finally, extensive experimental results demonstrated that HocSLM achieves state-of-the-art performance on three benchmark datasets: NTU RGB+D 60, NTU RGB+D 120, and Northwestern-UCLA. Future research will focus on integrating additional modalities into HocSLM and further optimizing its performance in real-time action recognition applications. 

\printcredits

\section*{Declaration of competing interest}
The authors declare that they have no known competing financial interests or personal relationships that could have appeared to influence the work reported in this paper.

\section*{Acknowledgments}
This work is partly supported by the Key Research and Development Program of Ningxia Hui Autonomous Region (2026BEE02004).

\section*{Data availability}
Data are publicly available.

\bibliographystyle{cas-model2-names}

\bibliography{cas-refs}

\end{document}